\documentclass[12pt,twocolumn]{article}

\usepackage[a4paper,
   bindingoffset=0.2in,
   left=0.75in,
   right=0.75in,
   top=0.75in,
   bottom=1in]{geometry}

\usepackage{sectsty}
\sectionfont{\large}
\subsectionfont{\normalsize}

\usepackage{microtype}     
\PassOptionsToPackage{warn}{textcomp} 
\usepackage{hyperref}
\usepackage{textcomp}     
\usepackage{mathptmx}     
\usepackage{times}      
\usepackage{cite}      
\usepackage{tabu}      
\usepackage{booktabs}     
\usepackage{overpic}
\usepackage{natbib}
\usepackage{adjustbox}
\usepackage{graphbox,graphicx}

\title{Is Medieval Distant Viewing Possible? : Extending and Enriching Annotation of Legacy Image Collections using Visual Analytics}

\author{Christofer Meinecke\footnote{Christofer Meinecke is with Leipzig University. cmeinecke@informatik.uni-leipzig.de.}, Estelle Guéville\footnote{Estelle Guéville is with Yale University. Estelle.gueville@yale.edu}, David Joseph Wrisley\footnote{David Joseph Wrisley is with New York University Abu Dhabi. djw12@nyu.edu.}, \\ and Stefan J\"anicke\footnote{Stefan J\"anicke is with University of Southern Denmark. stjaenicke@imada.sdu.dk.}}
\date{}

\begin{document}
\maketitle

\graphicspath{{figures/}{pictures/}{images/}{./}} 

\abstract{
Distant viewing approaches have typically used image datasets close to the contemporary image data used to train machine learning models. To work with images from other historical periods requires expert annotated data, and the quality of labels is crucial for the quality of results. Especially when working with cultural heritage collections that contain myriad uncertainties, annotating data, or re-annotating, legacy data is an arduous task. In this paper, we describe working with two pre-annotated sets of medieval manuscript images that exhibit conflicting and overlapping metadata. Since a manual reconciliation of the two legacy ontologies would be very expensive, we aim (1) to create a more uniform set of descriptive labels to serve as a "bridge" in the combined dataset, and (2) to establish a high quality hierarchical classification that can be used as a valuable input for subsequent supervised machine learning. To achieve these goals, we developed visualization and interaction mechanisms, enabling medievalists to combine, regularize and extend the vocabulary used to describe these, and other cognate, image datasets. The visual interfaces provide experts an overview of relationships in the data going beyond the sum total of the metadata. Word and image embeddings as well as co-occurrences of labels across the datasets, enable batch re-annotation of images, recommendation of label candidates and support composing a hierarchical classification of labels. 
} 


\section{Introduction}
Visual material stored in public collections is not always in shape to be used for state-of-the-art computer vision methods; that is to say, these collections are not ready as data. There are many reasons for this: the variable quality of the digital twin, limited availability of metadata, or uneven digitization due to the preservation status of the cultural object. In today's digital libraries, new digitization of manuscripts does not focus on the visual content, but on the codex as object. Paradoxically, while providing more overall content, they provide less metadata about images than the older iconographic databases. Researchers turn to the databases created in the era of partial digitization of the 1980s and 1990s for the rich metadata they contain.

These annotations in earlier cultural heritage collections provide a propitious opportunity for applying supervised machine learning methods. It is not a straightforward task, however, to use these annotated images for machine learning. Not only have the original vocabularies of various collections “drifted apart”, but the collections also contain manifold uncertainties, such as imprecision, incompleteness, and non-homogeneity in both the annotated metadata and in the data of the collection itself~\citep{borner2019network}. One interest lies in organizing and structuring similar datasets with overlapping metadata that allow researchers to gain deeper insight into specific materials or phenomena, which for artificial reasons have been siloed in different cultural institutions. Additionally, more robust general connections between divergent image collections should allow greater retrieval and discoverability in the cultural heritage sector between varied vocabularies or even across multilingual metadata schemas~\citep{angjeli2008semantic,gehrke2015biblissima}. This can even be used as a starting point for distant viewing of a large amount of medieval visual content.

Another problem with legacy annotations is that of missing labels or inappropriate hierarchies that do not match those usually used for machine learning. The assumption that labels are independent of each other often does not represent real-world scenarios, therefore, taking into account the relations between labels can improve classification~\citep{dhall2020hierarchical} or retrieval tasks~\citep{barz2019hierarchy}. The most common label hierarchy in computer vision is ImageNet~\citep{imagenet_cvpr09}, which uses a subset of WordNet~\citep{miller1995wordnet} labels. 
This hierarchy can be problematic for multiple reasons since its images and attendant hierarchies are not applicable to historical image domains. Also, relations in hierarchies like WordNet age poorly, e.g., outdated or offensive language, or they can include non-visual concepts problematic for image annotation~\citep{yang2020towards}. The problem of non-visual concepts for image annotation belongs not solely to contemporary image datasets but also can be found within historical collections, as in the case of the interpretation of narrative.

We designed visualizations as part of our research thinking process to combine two (partially-)annotated image datasets representing the same genre, but originating in two different research initiatives exhibiting differences and inconsistencies in their vocabulary. The visualizations were used for labeling purposes and provided the context for discussing the underlying domain situation. In particular, we unify the labels of a subset of the Mandragore and Initiale datasets by creating a shared, high-quality label hierarchy. We decided to construct the label hierarchy based on pre-existing labels as well as entirely new terms. A specific vocabulary related to the cultural horizons of the period was essential, and existing external hierarchies do not include the requisite period-specific vocabulary. For us, labeling means both assigning categorical labels to an image and defining relations between labels. We focus on labeling multiple images at the same time while suggesting existing labels from the different datasets. Recommendations are made using word embeddings, co-occurrences of the labels across the datasets, and image embeddings of the images in the collection.

Continuing our long-standing interdisciplinary collaboration~\citep{janicke2017interactive,meinecke2021explaining}, we adopted a participatory design process~\citep{gappaper} to address the above-mentioned issues. Our contributions can be summarized as follows:

\begin{itemize}
 \item A \textbf{multi-layered visual analytics process} that tailors Shneiderman's Information Seeking Mantra~\citep{shneiderman1996eyes} to navigate large sets of images from embeddings to detailed annotation views.
 \item A \textbf{multi-view image (re)-annotation environment} that provides various visual interfaces to explore various aspects of the data to help evaluate the similarity and relatedness of images.
 \item \textbf{Usage scenarios} documenting various strategies for how domain experts can use such visualizations to gain insight into their visual material.
 \item A \textbf{label hierarchy} for medieval illuminations produced by content specialists using the system. It can be straightforwardly applied to scenarios in which hierarchical classification or weakly supervised object detection~\citep{inoue2018cross} is performed on specific historical sets of images with related themes.
\end{itemize}

Although the domain of medieval manuscripts might seem quite specialized, the situation of divergent common vocabularies and the desire to resolve and combine labels across knowledge bases is common to many research fields. Our system is designed to support subject specialists from different backgrounds in viewing their sources (here, specialists in pre-modern culture, such as paleographers, art historians, codicologists, and philologists), and by extension, different publics in visual cultural studies which focus on different details in image corpora. Our solution is adaptable to related image annotation scenarios for the revision, creation, and/or organization of domain-specific labels in a hierarchical structure.

\section{The Complexity of Medieval \\Images and Visual (meta)data}

\subsection{Distant Viewing of Medieval Manuscripts}
Distant viewing in the context of medieval manuscripts, by which we mean the use of computational methods to explore a large number of digitized images of manuscript illuminations, is not a straightforward process. Most distant viewing research has been done using neural networks trained on twentieth- and twenty-first century imagery~\citep{wevers2020visual, arnold2019distant}. While it is true that photographs are old, they are still quite close to contemporary image datasets used for pre-training models. In this article, we use concepts and methods from visual analytics to explore if it is also possible to extend and enrich annotation of legacy image collections. 

We use a corpus of images from thirteenth- and fourteenth-century Latin bibles, exhibiting a certain degree of uniformity in their layout; among other details, one finds decorated initials at the beginning of chapters or prologues, intercolumn decoration and decorative outgrowths. These commonalities suggest that predictable patterns in the kind of objects, images, colors and decorative elements are present in these bibles. For example, the book of Judith might be decorated with a number of different depictions of the core narreme of the book, the beheading of Holofernes, which point to different traditions of representing this story in time and space. An analogous question exists for the textual content of our corpus, predicting patterns in the use of abbreviations or inclusion of prologues, but the study of textual variance goes beyond the scope of this article ~\citep{gueville2022transcribing}.

We study the coherence of the visual tradition of the corpus using the tools of computer vision, and in doing so we aim to put together new building blocks rooted in medieval imagery for further research. We also believe that computer vision applications in corpora of medieval illumination have great promise, particularly since specialists in iconography generally agree that images exist in series with a significant coherence over time and space~\citep{baschet}. Computer vision research applied to medieval manuscripts has been carried out before: to spot patterns~\citep{ubeda2020improving}, to classify crowns~\citep{yarlagadda2010recognition} or gestures ~\citep{schlecht2011detecting} as well as more generally to organize medieval manuscripts in the service of iconographic research~\cite{lang2018attesting}. Other work has aligned images in manuscripts using collation methods~\citep{kaoua2021image}, or aligned text and image data of medieval manuscripts with visual-semantic embeddings~\citep{baraldi2018aligning, cornia2020explaining}. Our approach's methodological intervention is to apply advanced interactive visualization methods, first, to explore the results, and second, to adjust and enrich collection metadata.

Arnold and Tilton~\citep{arnold2023distant} argue that the project of distant viewing is not about the brute application of computer vision algorithms to image datasets but rather it must design exploratory processes that "mirror the approaches we turn to when working with a smaller collection of images". 
Furthermore, it must also take into consideration both how visual materials make and transmit meaning, and how computer vision algorithms "mediate the interpretation of digital images". 
Some features of medieval images, we have found, are relatively easy to detect computationally, for example, people, animals, or trees. This is no doubt due to the presence of similar objects in contemporary image hierarchies. Thanks to labeling in legacy image collections, we trust that it will also be possible to identify other highly specialized objects such as crowns or swords, uncommon or complex varieties of flora or fauna, or even the hybrid human-animal figures or imaginary animals so common in medieval marginalia~\citep{camille1992image}. Medieval iconography is also interested in body postures and gestures, which we believe, with enough training data, are also computer "seeable". 

\begin{figure}[t]
 \centering
 \includegraphics[width=0.7\linewidth]{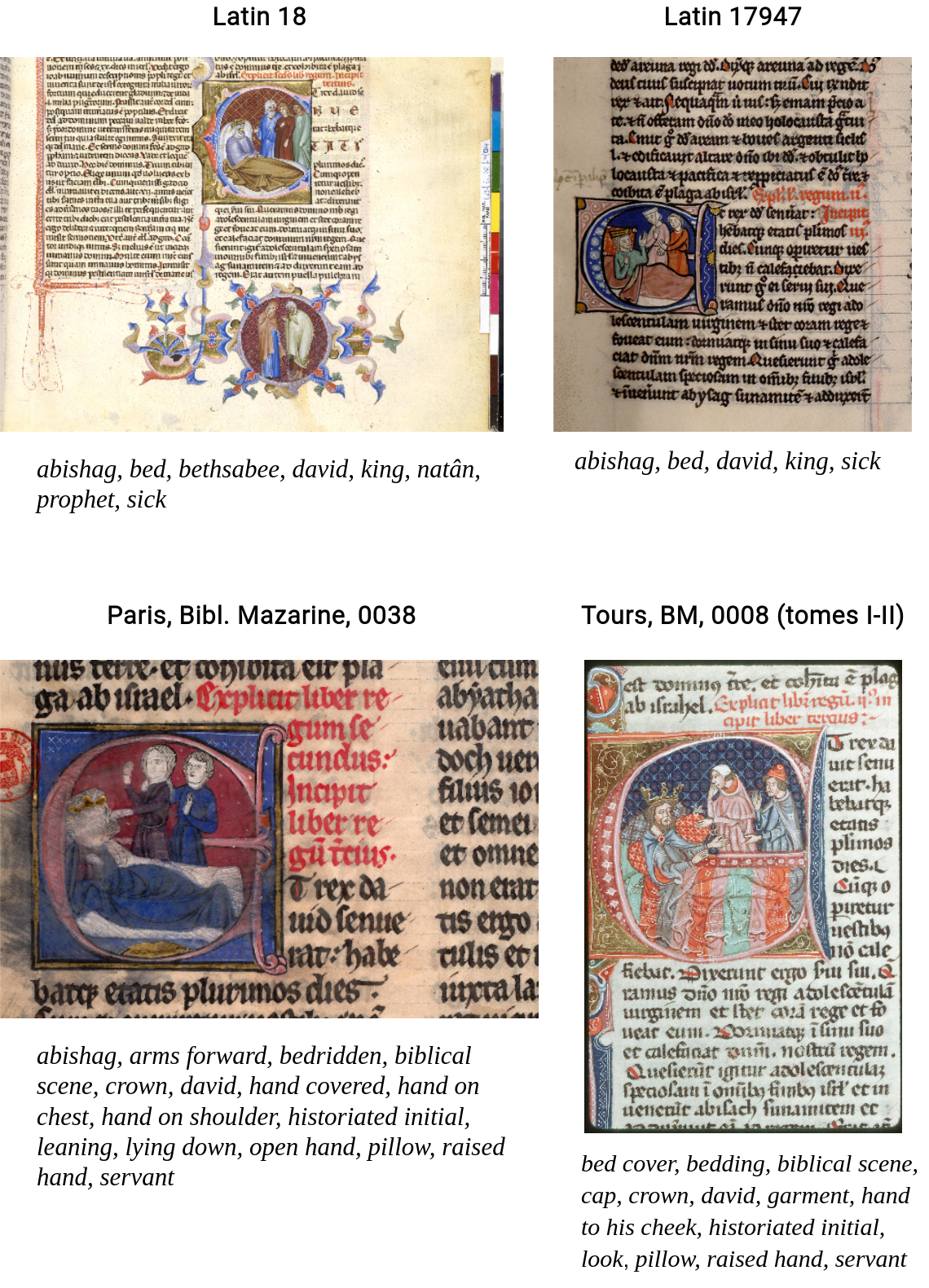}
 \caption{Examples taken from 1 Kings in the dataset with their labels. The upper ones (from Mandragore), and the lower ones (from Initiale) illustrate possible variation of the images in the dataset. Even in these cases where the same scene is depicted, preservation status and the background colors vary. They also show how Mandragore and Initiale focused on different concepts in the images with the inclusion of positions and gestures in Initiale. Depicted here are four manuscripts: Bibliothèque nationale de France, Latin 18, folio 104r and Latin 17947, folio 112r, Bibliothèque Mazarine 38, folio 230v, and Tours, Bibliothèque municipale 8, folio 230r.} 
 \label{fig:data}
\end{figure}

Unlike most photographs, medieval images may include objects that connote something other than the literal meaning, as in the case of the key that Saint Peter is usually carrying, a reference to the keys of the kingdom of Heaven referenced in the book of Matthew. Other parts of images often make meaning in an allegorical or a narrative way. That is to say, extra-visual information is needed to contextualize and interpret the image. Expert annotators may label an image as being about the "Creation" or "King David," in the absence of any specific visual cues that transmit such a meaning; it is rather the narrative context of the biblical text that sits behind these labels. In \autoref{fig:data} (below), labels for "David," "Abishag," "bed," and "sick" are associated with a set of Paris bible illuminations, combining direct and indirect, contextual meanings for a scene of the old age of King David depicted in 1 Kings. Only the bed could likely be detected by current computer vision algorithms.

Decorative patterns recur through the manuscripts of legacy databases, sometimes with remarkable similarity across time and space. Illuminators, whose work it was to decorate bibles (often asynchronously with the copying of the text itself) were not simply copying images from exemplars, but were drawing upon a larger tradition of representation, a cultural memory of potential ways of visually interpreting the textual source. Indeed, one story from the Bible can be represented in any number of ways, with patterns of inclusion or exclusion of iconographic details. This kind of visual variance has been understood as a meaningful interpretive layer in manuscripts, and cataloging systems have tried, to a certain extent, to reflect that variance by using thesauri to describe these images.

\subsection{Complexities of Legacy Databases}

Up until the early thirteenth century in Europe, manuscripts were mostly produced in workshops attached to courts or by monks in monasteries, but the creation of universities greatly influenced this form of written, cultural production. "Paris bibles", a tradition that we have called elsewhere one of variance in uniformity, emerged in thirteenth century Europe as a mass-produced written object in response to new forms of literacy, namely teaching and preaching~\citep{light2012}. After 1220, these hand-copied Bibles contained a corrected text and followed a standard order, introduced by prologues and divided into chapters, usually including related series of illuminations and decoration collocated with the prologues and chapter beginnings. At first glance, the images found in Paris bibles seem quite similar from one manuscript to another, yet closer examination allows us point out differences: the use of colors and details of representation, the presence or absence of objects or people, differences usually attributed to the origin of the manuscript, to the individual illuminator or workshop. 

Our dataset of images from Paris bibles contains a subset of two databases: Mandragore~\citep{mandragore} and Initiale~\citep{initiale}; with respectively 1633 images from 53 manuscripts and 11472 images from 241 manuscripts. Each digital image illustrates one or two pages of a manuscript and can contain one or several illuminations. In addition to the images, the dataset also includes general spatial and temporal information about the manuscripts, a topical description of the images, and labels indicating the book of the Bible depicted. Both databases include descriptive details that can be recognized without expert knowledge such as objects (table, seat, sword, etc.) or body positions – as well as decorative and interpretive details that can be understood with specialized iconographic knowledge and the context of the textual tradition. There are also many records that have codicological metadata, but for which no labels have been created.

Mandragore is an iconographic database~\citep{aniel1992mandragore} created at the Bibliothèque nationale de France in 1989 and made available to the public in 2003. It describes the general iconographic features of more than 200000 illuminations, drawings and decors~\citep{mandragore_new}. The controlled vocabulary it uses was originally based on the "Thésaurus Garnier"~\citep{garnier1984thesaurus} but expanded beyond it. It now includes about 21000 unique labels, 530 of which are used in this subset. The database continues to be enriched by curators, librarians, and researchers with the aim of identifying, in each illumination, all the objects, places, people, and iconographic subjects represented. A new web-based version was published in November 2022, which includes IIIF manifest and a Mirador viewer. Additionally, most of the labels, except names, are now translated into English, German, Italian and Spanish~\citep{mandragore_new}.

Created in the early 1990s at the Institut de recherche et d'histoire des textes du Centre national de la recherche scientifique - Section des manuscrits enluminés, Initiale is a web-based catalog of medieval manuscripts belonging to the public libraries of France, excluding the Bibliothèque nationale de France. It includes about 10000 manuscripts with more than 90000 illuminations. With specialized iconographic research in mind, it uses a refined iconographic index with a controlled vocabulary and offers specialized, art historical analysis of the decoration. This vocabulary was also originally based on the 
"Thésaurus Garnier"\citep{garnier1984thesaurus} but it has evolved over the last 30 years~\citep{lalou2001base}. Our subset uses 1734 words from this controlled vocabulary, only 279 of which are shared with Mandragore.

Although both databases contain samples of Paris bibles, they were created at very different times and with different priorities. While deep, the divide between Mandragore and Initiale is an artificial one, owing to the history of institutions and collections, rather than the original historical material. 

This situation is far from optimal for scholars who would like to see relationships in the larger picture of medieval manuscripts.
Collections as data for computational analysis were not on the minds of their creators. A question that informs our work is to what extent are these legacy databases reusable for purposes other than the ones they were originally created? 
One of the objectives of our work is to bridge the gap between the different databases by combining, regularizing and extending the two vocabularies so that the metadata describing the images can be used and analyzed together. We do not aim to create fuller descriptions of the scenes depicted by hand but rather want to make sure the vocabulary is inclusive, organized and detailed enough for the computer to understand.

\section{Visual Thinking Process}\label{s:3}
Visual analytics has been defined as the combination of “automated analysis techniques with interactive visualizations for an effective understanding, reasoning, and decision-making on the basis of very large and complex datasets''~\citep{keim2008visual}. This allows us to explore domain-specific research questions by using interactive visualizations together with (semi-)automatic methods such as computer vision or classical machine learning. Traditionally, this leads to a strong focus on the development of systems that should help in answering domain-specific research questions. The purpose of visualization is not only to present data in the form of a system or tool that helps in performing a specific set of tasks but also to serve as a research thinking process and a mediator~\citep{hinrichs}. In the case of datasets of more than 10000 medieval illuminations, one view is most certainly not enough. Rather, filtered and aligned views that incorporate both traditional metadata and computed results designed in collaboration with subject specialists contribute to thinking through the research problems. Hinrichs et al.~\citep{hinrichs} argue that visualization can also be used as a speculative process through the large design space of potential representations and visual channels. In line with this notion of visualization, we engaged in repeated, long discussions about what aspects of the underlying data to focus on, how they can be aggregated and displayed, as well as how we can make use of the large set of already existing labels associated with this large image collection.

Our resulting visualizations help with the specific tasks and research questions, but naturally, they need to evolve or be replaced depending on the development of the ongoing cooperation, as humanities research questions can be developing, complex, open-ended and advanced by serendipitous discoveries~\cite{thudt2012bohemian, liestman1992chance}. The interactive component of visualizations provides scholars with new means to interact with their research material; similarly, the visual component also helps to see patterns and relationships that were previously hidden in the collection. This is especially true when engaging with visual research material. Nonetheless, in the digital humanities, the use of singular or static visualizations without many interaction mechanisms is quite common. Often visualizations focus only on a singular aspect of the underlying cultural collection. But for most cultural collections their complexity cannot be addressed by just one standalone visualization~\citep{dork2017one}. 

Indeed, working with large collections of cultural material is not always a linear process. This is especially true for data that were previously structured and labeled by several people with different goals in mind, since such data can have inconsistencies and contradictions that are not always visible to the researcher. By exploring the different facets of the data we can revisit previous decisions and understand the intentions of the previous labelers. If we admit that visualization is part and parcel of the thinking process, it is not a surprise that a considerable number of visualizations that were created were then subsequently dismissed as “sandcastles,” that is, they helped in the iterative thinking process but were ultimately not included in the workflow. Visualization allowed us to uncover the complexities of data created over time and with many different agendas in mind, and it also helped us understand how we might go about bridging the vocabularies through different forms of normalization and to formulate new research questions that go beyond those enabled by the original database.

\section{Processing of Source Material \\\& Visual Labeling}
\subsection{Data Processing}
The first computational step consists of processing images, textual metadata, and labels. For image pre-processing, we tested multiple methods to extract the illuminations from the images. Although the method in Grana et al.~\citep{grana2011automatic} based on the Otsu algorithm~\citep{otsu1979threshold} showed good results on subsets of the dataset, it was not adaptable to the entire corpus due to the varying preservation status and background colors of the manuscripts. Furthermore, we tested pre-trained models of the docExtractor~\citep{monnier2020docextractor} but the results were not satisfactory for the entire corpus. So, we used the original image data from Mandragore and Initiale. 

For the next image processing steps, we apply the EfficientNet B7~\citep{tan2019efficientnet} that was pre-trained on ImageNet ~\citep{imagenet_cvpr09}. We use the top layer of the network to compute the image embeddings for each image in the dataset. These embeddings are used to compute nearest neighbor similarities between the images. Although the network was trained on natural images, the features can still be used to compute similarities between images in the corpus~\citep{crowley2014search}.

All embeddings are added to a faiss~\citep{johnson2019billion} index structure based on the Euclidean distance between them. For each image, the most similar images can be queried by their embedding. Pre-processing of image data in the form of object detection would be valuable; however, high-quality label hierarchies describing medieval illuminations do not exist, and object detection based on modern hierarchies like Open Images~\citep{kuznetsova2018open} fail. Few objects depicted inside our image dataset are found, since many classes of these hierarchies did not exist in the historical past.

For the pre-processing of the labels, we lowercase all words, remove diacritics and special characters, followed by the application of two pre-trained models for modern French fastText~\citep{bojanowski2017enriching,grave2018learning} and CamemBERT~\citep{martin-etal-2020-camembert} to embed the labels. For the fastText model, we use word vectors, and in cases where a label is composed of multiple words, we compute an average vector. For the CamemBERT model, we apply mean pooling to the hidden state embeddings of the neural network. We did not include stopwords from labels with more than one word to prevent high similarity in cases where stopwords overlap. Following this, we add the vectors to a faiss index. Some images have descriptive sentences for which we also compute embeddings. The fastText vectors appeared to grasp the word relations better based on the nearest neighbors of the labels. Since most labels are single words, we disregarded the CamemBERT embeddings, although they could be better in other application scenarios where multiple sentences are used.

We also compute for each image in the dataset three types of two-dimensional embeddings with UMAP~\citep{mcinnes2018umap-software,mcinnes2018umap}, which we use to visualize the images in a two-dimensional space. The different UMAP embeddings are based on the image embeddings, the image label's word embeddings, and the description's embeddings. We use a fixed random state for the computation. This is important for recomputation after labeling by the domain expert. 

To display similarities between manuscripts, we compute multiple measurements. The similarities are defined by the Euclidean distance between average vectors. For image similarity, we use the image embeddings of all images belonging to a manuscript. Label and description similarity are defined in a similar way, by using average word vectors of labels and descriptions associated with a manuscript's images. Combining image similarity with label or description similarity can be helpful: similar images depict similar scenes, and similar scenes are likely to have similar labels or similar descriptions. Since not all images in the dataset are labeled, the image similarity allows us to include them.

\subsection{Visual \& Interactive Data Labeling}

\begin{figure*}[h]
 \centering
 \includegraphics[width=\linewidth]{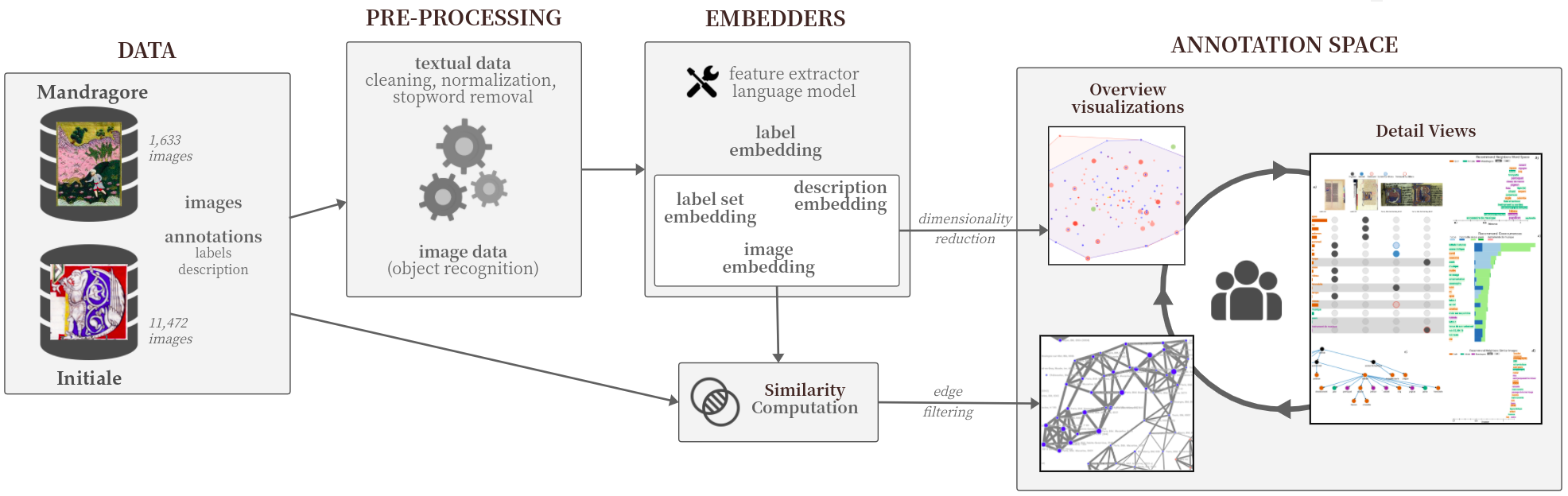}
 \caption{Systematic overview of our image and label exploration and annotation workflow.}
 \label{fig:overview}
\end{figure*}
The focus of our work lies on image labeling in combination with distant viewing, which is similar to visual annotation systems, interactive labeling and other human-in-the-loop processes. In order to label visual material it is important to see the object of interest, the associated metadata and the possible relations between them. For example, the ability to annotate spatial regions in images is required for the application of localization methods, such as object detection. Annotation tools like the VGG Image Annotator~\citep{dutta2019via}, CVAT~\cite{CVAT}, or EXACT~\cite{marzahl2021exact} support this but without visualizations to communicate features of the data. Although annotating spatial regions like bounding boxes is often important, it is not the focus of our work at this stage, as we first want to regularize the vocabulary of the already existing labels.

In general, visual annotation systems support manual labeling tasks through visualizations. Some systems support only a small predefined vocabulary~\citep{willett2011commentspace} or properties of the data like outliers~\citep{chen2010click2annotate}.
While others support any textual annotations without a defined vocabulary~\citep{elias2012annotating}. In contrast, we use a pre-existing vocabulary, extending and regularizing it. Other systems focus more on collaboration through monitoring functionalities like inter-annotator agreement~\citep{qwaider2017find}. As we currently designed the system for a small number of experts, we did not focus on the collaboration aspect in detail, instead, we show the labels and images that another user worked on. 
Furthermore, labeling can be supported by visualizations to explore the data, as well as machine learning methods to recommend points of interest~\citep{felix2018exploratory, khayat2019vassl} contributing to the sense-making process~\citep{endert2017state} and building trust\citep{chatzimparmpas2020state} in the machine learning model. 

Modern visual analytics systems combine interactive visualizations with active learning strategies~\citep{wu2020multi} to understand the classifier better and to support tasks such as correcting mislabeled training data~\citep{xiang2019interactive}, annotating text data~\citep{kucher2017active, sun2017label}, labeling documents~\citep{choi2019aila}, constructing sentiment lexicons \citep{makki2014context} or classifying the relevance of tweets in real-time~\citep{snyder2019interactive}. 
~\cite{bernard2018vial} introduced visual interactive labeling (VIAL) as a method for combining active learning with visualization systems for the exploration and selection of data points for labeling. Visual encodings that expose the internal state of the learning model can help in the labeling process~\citep{liu2018perceptual}.

Our approach shares similarities with works on visual-interactive labeling. We do not include active learning strategies, but our exploratory approach can be extended to encompass recommendations based on active learning. Currently, the identification of labeling candidates is a choice made by expert users. They have visualizations at their disposal displaying similarities based on embeddings and metadata. We exclude active learning at present, for a few reasons: the large vocabulary size (exceeding 2000 labels), the skewed distribution of the existing labels, and the multi-label classification setting. Hierarchical classification helps to mitigate these circumstances.

\section{Visual Analytics for \\Medieval Manuscripts}\label{s:5}
We visualize images of cultural heritage~\citep{windhager2018visualization}, but we combine exploration analysis~\citep{thudt2012bohemian, dumas2014artvis,dork2017one, jocch} with image labeling. For this, we created several visualizations as entry points to the data, in order to engage in an analysis of the images and their labels. 
The visualization sections include usage scenarios by the two medievalists that worked with the system.
A systematic overview of our workflow can be seen in \autoref{fig:overview}.

After the data processing, we aggregate the embeddings by manuscripts to which they belong. The inter-manuscript relations are then visualized in a graph. We use a point cloud visualizing each image based on two-dimensional embeddings to present and filter images. Then, we enable the inspection and selection of a specific subset of the image data and their labels on the basis of the same or similar metadata or relations in a vector space. This is needed to allow one to annotate images by adding missing concepts, or by objects that are depicted, but unlabeled. Different recommendation methods can support this task: word and image similarities, co-occurrences, or even active learning methods. The (re)-annotation space also includes an interactive graph to support creating a high-quality label hierarchy for medieval illuminations.

\subsection{Manuscript Graph}\label{s:5.1}
\begin{figure}[t]
 \centering
 \includegraphics[width=\linewidth]{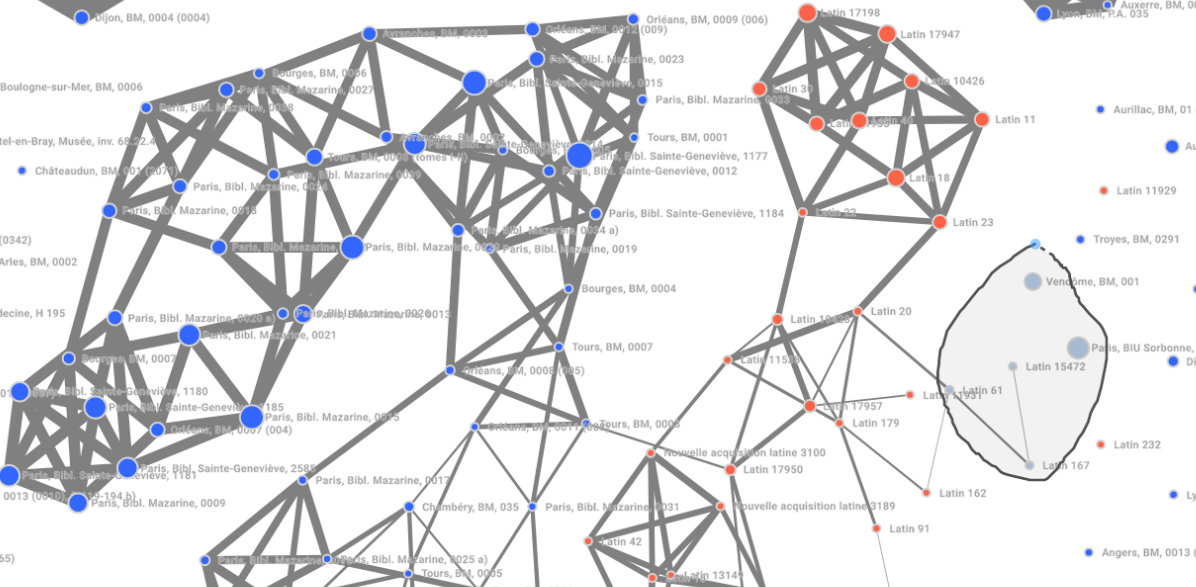}
 \caption{An excerpt of the manuscript graph based on label similarity. Blue nodes are part of Initiale and red nodes are part of Mandragore. Showing the separation of both datasets and the similarity between the manuscripts in the respective dataset. The grey area illustrates the user-selected manuscripts, including some of them that have overlapping labels (connected) and others that do not (without connection).}
 \label{fig:manu-graph}
\end{figure}

The manuscript graph (\autoref{fig:manu-graph}) displays manuscripts as nodes and connects them by edges based on user-selected similarities. This visualization serves as the entry point into the collection, filtering manuscripts of interest so that their images can be accessed. The nodes are color-coded according to the dataset to which they belong, and the size indicates the number of images that belong to this manuscript. It is a force-directed graph in which the nodes repel each other and the edges pull the nodes together. The edge thickness shows the selected similarity value.

The domain expert can select one or multiple similarity metrics for the graph, such as image similarity, label similarity, or description similarity. When two or more similarities are combined, the edge value corresponds to the average of the values. To avoid visual clutter, it is possible to select the maximum number of edges a node can have and filter all edges below the chosen threshold. Additional similarity metrics can be added as a graph overlay to assess its impact on the graph. It is also possible to drag nodes to another position and zoom in and pan the graph. In order to explore a specific subset of manuscripts, one can use a lasso selection to draw a polygon around the nodes. The selected nodes are then increased in size and colored steel-blue. A drawer on the left side can be toggled, showing a bar chart, a timeline, and a label cloud with information on the metadata of the selected manuscripts.

\begin{figure*}[t]
 \centering
 \begin{overpic}[trim={0.365cm 0 0 0},clip,width=.49\linewidth]{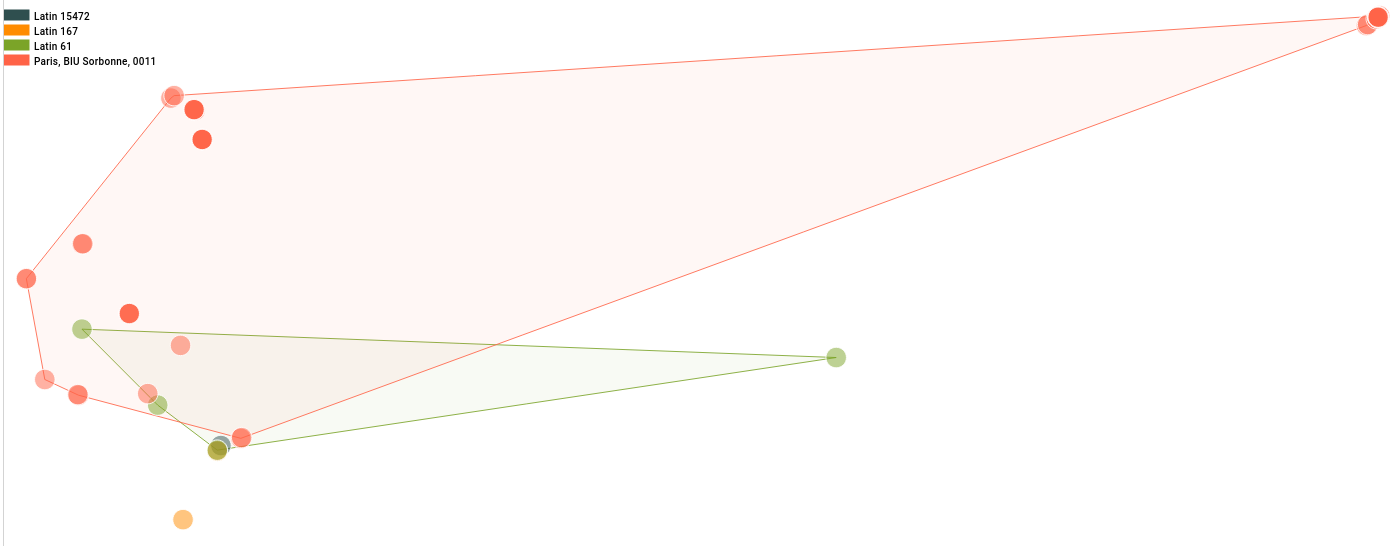}
 \put(1,28){\textbf{\textit{a)}}}
 \end{overpic}
 \begin{overpic}[trim={0.163cm 0 0 0},clip,width=.49\linewidth]{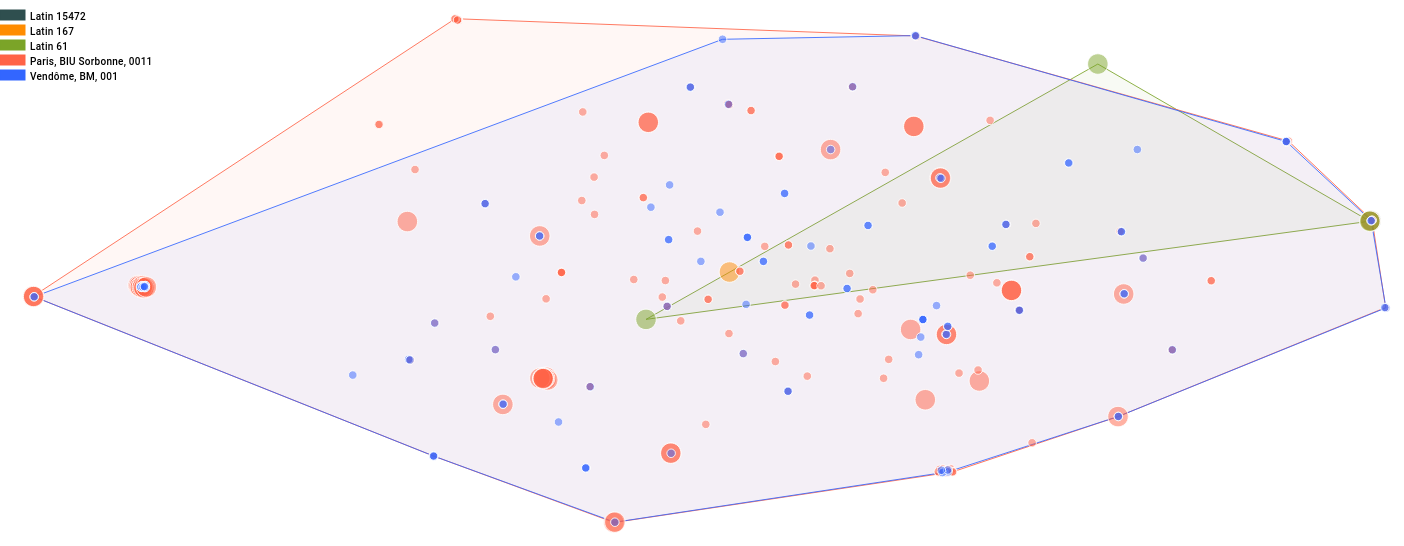}
 \put(1,26){\textbf{\textit{b)}}}
 \end{overpic}
 \caption{A point cloud of images based on the word embeddings of the labels (a), where only images with labels are visible. After selecting some of the images and changing the used embeddings to the textual description (b) several images without labels are displayed next to or on top of already labeled images. Notably, several images of Vendôme, Bibliothèque municipale 1 have almost the same description as some of the already labeled images. This visual design allows for finding sets of images with the same or similar content amongst a set of labeled and unlabeled ones.}
 \label{fig:point-cloud}
\end{figure*}

\textbf{Usage Scenario.}
The basic idea of this visualization is to be able to choose any subset of manuscripts in the dataset for further work. One approach to the lasso selection is to connect manuscripts from the two datasets, the idea being to attempt to bridge the gap between the two label vocabularies. As one might expect, a small overlap between the two vocabularies exists, but for the most part, the two phases of legacy annotation focused on different aspects of the data. The calculated overlays provided additional metrics for assessing similar manuscripts, although we did not explore these to the fullest extent. 
Guéville, Estelle, and Wrisley, David Joseph. “Transcribing Medieval Manuscripts for Machine Learning.” Journal of Data Mining and Digital Humanities. 2023
Another strategy at this stage would be to avoid focusing on already linked manuscripts. This prevents reinforcing the existing connections. Instead, we would select a group with a trade-off of connected and unconnected points. Such an example can be seen in \autoref{fig:manu-graph}. The two medievalists took different approaches. One would select a smaller number of manuscripts to work on at a time, whereas the other would select a larger number, and then apply subsequent filters to limit the results, for example, by the book of the Bible. The former was focused on overall similarities of the manuscripts (location, dating), whereas, given the nature of the iconography of Paris bibles, the latter tended to focus on variations on a visual theme. Whereas the risk of one approach was not finding many connections; the payoff of the other was a greater benefit of finding new connections. In general, the manuscript graph was quite useful for asserting broad relationships based on existing metadata. 

For each selection of manuscripts, a visual sidebar allowed us to see inventory numbers, total number of images per manuscript, as well as spatio-temporal metadata. While an interesting feature, the promise of being able to connect labeling to specific times and places in order to see patterns was more useful in the case of already connected manuscripts. Overall, this approach was challenging, no doubt due to the lack of granularity and precision of those features. We would expect this feature to be more useful the more connected the datasets will grow to be. 

\subsection{Image Point Cloud}
The Image Point Cloud (\autoref{fig:point-cloud})is an entry point to the labeling process by showing two-dimensional representations of the images of the selected manuscripts so that similar images are presented close to each other in the space. In it, each circle represents an image. One can select and combine embeddings based on the images, the labels, and the description of the images. 
 
Each selected manuscript has a different color, which is displayed in a legend together with the manuscript name. To highlight the positions of the images of a specific manuscript, the convex hull of the points is drawn as a contour. It can be toggled by clicking on the manuscript in the legend, and it is possible to zoom in and pan the point cloud. In the case of similar or the same two-dimensional representation for multiple images, or when multiple manuscripts with a large number of images are selected, overplotting can result. Filtering displayed points based on metadata, or drawing a rectangle around them to recompute the layout are two possible solutions, allowing us to focus on a specific subset of images for labeling.

On mouseover, the image is shown. A lasso selection can be used at the points to select a set of images for the labeling process. The selected points are increased in size to better highlight them. The re-annotation space is accessed from a button in the left-hand drawer. The current state of the graph and the point cloud are saved at this point for returning to that stage of the labeling.

\textbf{Usage Scenario.}
Using description similarities or image similarity, it is possible to focus on the images directly in the Image Point Cloud. Four ways to work with the data were included, all of which could employ the lasso selection method. If a small number of manuscripts as described in Section~\ref{s:5.1} had been chosen, no filter would necessarily be required to begin to annotate. In the case of a large number of manuscripts in Section~\ref{s:5.1}, one could also not filter, but at the risk of having far too many examples in later steps to work with effectively. 

Of the possible filters at this stage (book, labels, subject), one can be used at a time. A filtered approach allows quite granular discovery of the images, their labels and the visual similarities. Each of the approaches was productive, with slight differences. Filtering by book provides the opportunity to explore visual content specific to the textual content of the biblical book. 
For example, for the book of Mark, one might find Mark depicted as a scribe sitting at a desk writing or in his more metaphorical depiction as a bull. 
The same image of a scribe sitting at a desk, for example, could exist across many different sections of the Bible (Jerome, Mark and the other Evangelists, etc.). 
Filtering by book or by subject is particularly useful for identifying visual variance within or across manuscripts, however a filter by book would be better since it would uncover examples of unlabeled images. 

Labels, on the other hand, allow us to delve quite further into the visual content of the bibles, even though we lose some of the specificity of narrative or subject. Examples of labels could include "pupitre" (writing desk), "chantre" (cantor), "hybride" (hybrid), "sommeil" (sleep) or "cigogne" (crane). Depending on the goals of the project, any of the four above-mentioned approaches can be selected before moving on to re-annotation.

\subsection{Re-Annotation Space} 
\begin{figure*}
\centering
\begin{minipage}{0.535\linewidth}
 \begin{overpic}[trim={0.2cm 0 0 0},clip,width=\columnwidth]{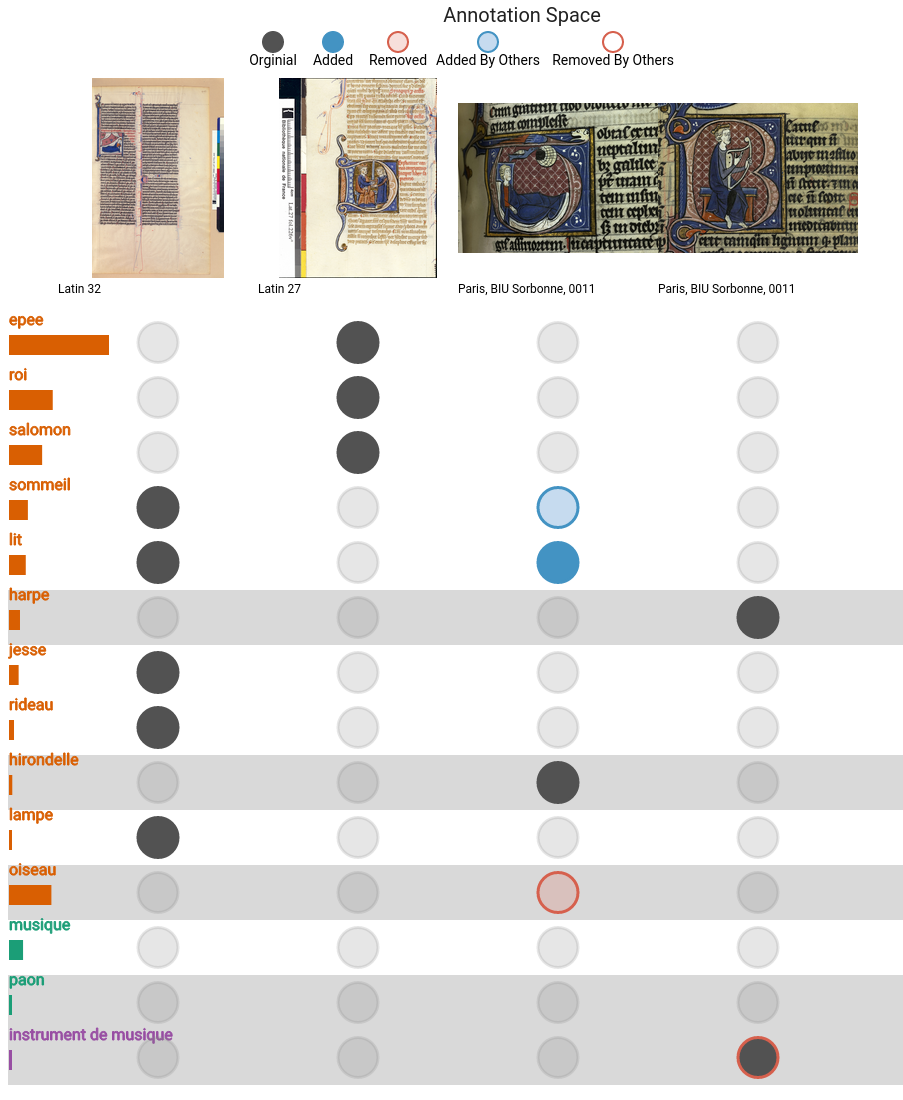}
 \put(1,90){\textbf{\textit{a)}}}
 \end{overpic}

 \vspace{1cm}
 
 \begin{overpic}[width=\columnwidth]{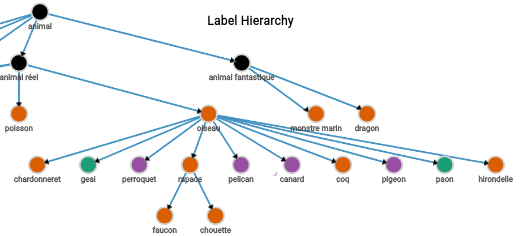}
 \put(70,40){\textbf{\textit{e)}}}
 \end{overpic}
\end{minipage}
\hfill
\begin{minipage}{0.335\linewidth}
 \begin{overpic}[trim={0 0 0 0},clip,width=\columnwidth]{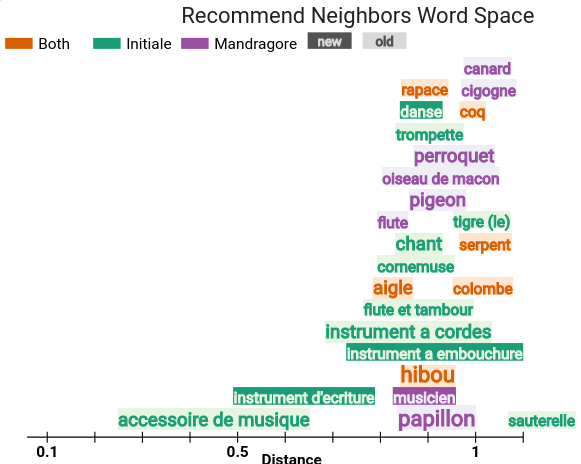}
 \put(99,75){\textbf{\textit{b)}}}
 \end{overpic}
 
 \vspace{0.5cm}
 
 \begin{overpic}[trim={0.2cm 0 0 0},clip,width=\columnwidth]{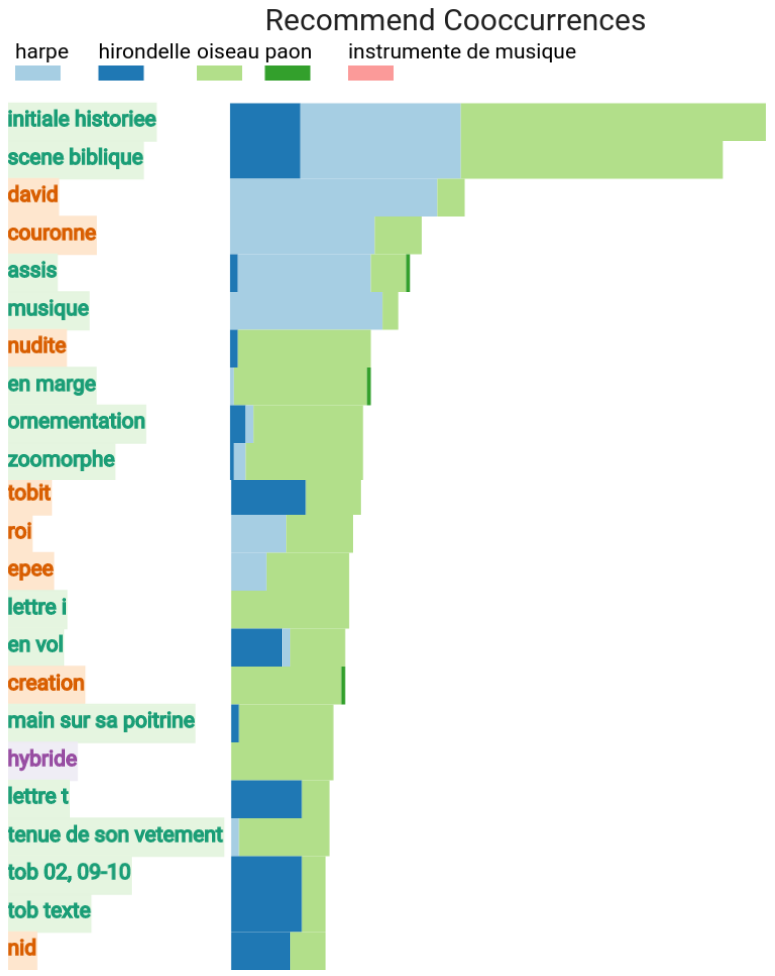}
 \put(80,95){\textbf{\textit{c)}}}
 \end{overpic}
 
 \vspace{0.5cm}
 
 \begin{overpic}[trim={0 0 0 0.1cm},clip,width=\columnwidth]{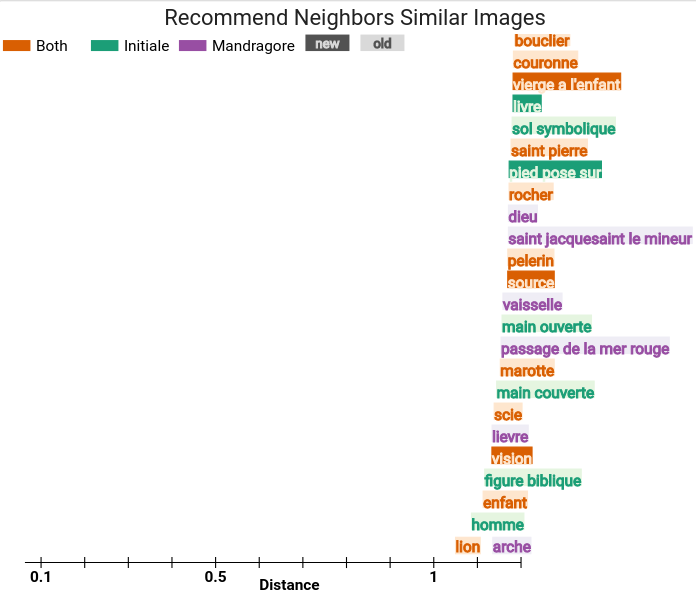}
 \put(95,81){\textbf{\textit{d)}}}
 \end{overpic}
\end{minipage}%
 \caption{The annotation space (a) shows four manuscripts and their labels arranged in columns. Some labels were added by different users; others were replaced by more specific ones. The word space (b) shows words that are similar to the ones currently selected in the annotation space. Here it is shown that after selecting "instrument de musique" (musical instrument) multiple words related to music were added. The recommended co-occurrences (c) show, for example, related terms such as "musique" (music) and "couronne" (crown), suggesting the relationship between music and King David. The recommended neighbors from the most similar images based on computed features (d) contain entries about animals. The excerpt of the label hierarchy (e) shows multiple labels of "oiseaux" (birds) asserted by the user. Manuscripts represented include Bibliothèque nationale de France Latin 32, folio 443r, and Latin 27, folio 226v, and Paris, Bibliothèque interuniversitaire de la Sorbonne 11, folio 150r and folio 169v.}
 \label{fig:annotation}
\end{figure*}
The annotation space (\autoref{fig:annotation}) allows one to see the current annotation status of a number of images as well as to add and remove labels and zoom in on the details of the images. It is also possible to add new labels that are not part of the dataset and to filter images based on metadata.

The selected subset of images is placed on the top and their position is fixed to allow them to be seen when scrolling through the list of labels. At left (\autoref{fig:annotation}a), the current labels of the images are presented alongside a bar chart illustrating frequency of appearance in the corpus, color-coded based on the dataset. When clicking on a bar, one can see other images that are annotated with this label in a pop-up. For each word-image pair, a circle is shown colored black for labeled cases or gray and less saturated for unlabeled ones.

Clicking on a circle, a label can be added or removed. All saved changes can be inspected in a history pop-up showing the timestamp, the user, and the changes. This allows annotators to keep track of their interactions as well as those of others. Images can be viewed in high resolution in a pop-up If a specific word is not of interest to the annotation process, it can be removed from the annotation space with a click. Visualizations are linked to the annotation space so that any update of the annotation space updates the other views. It is possible to select one or multiple words for the other views.

At right are recommended labels(\autoref{fig:annotation}b). The first visualization displays the words most similar to the currently selected words in the annotation space, similar to the word space view in the iteal system in~\citep{meinecke2021explaining}. Each word is placed on the x-axis depending on its minimum distance from the target words. The words are colored according to the corresponding dataset. We use a less saturated color to present either the background for old words or the font color for new words. When hovering over a word, we use a tooltip with the words in the annotation space that are the most similar to explain its recommendation. We use the same visualization to show the most similar labels of the most similar images in the current selection (\autoref{fig:annotation}d). This can be helpful in cases where the selected images do not have labels, but similar images do.

In \autoref{fig:annotation}c the most frequently co-occuring words with the currently selected words are displayed, ordered by their total number of co-occurrences. To distinguish between the co-occurrences with different words, we use a stacked bar chart for each word, ordered on the position in the annotation space. Because the number of labels can grow fast, the colors repeat themselves every 12 steps. However, this is not a problem when selecting a small number of labels in the annotation space to get recommendations. On hovering over a rectangle, the number of co-occurrences is displayed in a tooltip. Similarly to the previous visualization, we use a less saturated color to either present the background for old words or the font color for new words. The recommendations and co-occurrences also help to construct a label hierarchy by showing related labels.

\textbf{Usage Scenario.}
The re-annotation portion of the visual analytics system allows for a synoptic visualization of the labels for relabeling. A variety of approaches can be adopted to choose the materials. In general, aligning multiple images together in the same (re-)annotation space raises interesting questions about how iconographic difference helps to improve labeling. Given the general nature of the illuminations in the Paris bibles, filtering by book at this stage is a useful way of paying attention to iconographic variance when there is a common narrative element. 

User 1 chose a "trade-off" set in the manuscript graph, split between ones that had been found to be similar and others that had not, as one way of examining the work of previous annotators. The selection of a small set of four to six images for annotation inevitably led to many false pathways and re-selection of new images, but in combination with the recommendations in the word space and the co-occurrences and guided by the color coding, helped identify numerous labels not included in either the Initiale or Mandragore datasets. 
In fact, when a missing label was identified, the recommended images provided an effective way to cycle through possible candidates for annotation. 
Important in this process in the initial rounds were the indications of frequency, which allowed for commonly occurring labels to be explored with priority. In the cases of synonymous or near synonymous labels, the hierarchy was useful in the beginning as a way of linking and establishing an order between them.

The space allows to progressively check and extend the metadata at will, as well as remove any extraneous labels. On the other hand, depending on the frequency of the labels or subjects that can be used to filter at the previous stage, the annotation space can become crowded with aligned images that do not exhibit relevant relationships, and re-annotation becomes unlikely.

\subsection{Label Hierarchy}\label{s:5.4}
The label hierarchy view shows the current state of the underlying label hierarchy. We use the Sugiyama framework~\citep{sugiyama1981methods} to draw a directed acyclic graph for the label hierarchy. In the first step, it is checked with a depth-first search for each node if the graph contains cycles. If there is one, the edge is removed from the hierarchy but added back after the layout is computed. In the next step, nodes are assigned a layer for which we use the Network Simplex method~\citep{gansner1993technique} to minimize the length of edges in the graph. Then the nodes are ordered on the layer they are assigned to reduce edge crossings. For this, we use a top-down one-layer crossing minimization approach. In the last step, the nodes are assigned a specific coordinate using the quadratic programming approach, while minimizing the distance between the connected nodes, the curvature of the edges, and the distance between the disconnected components. 
After the layout is computed, the nodes are color-coded based on the dataset to which they belong. An excerpt of the hierarchy can be seen in \autoref{fig:annotation}e. The edges removed from the cycle detection are drawn in red to indicate to the domain expert that there is a conflict to be resolved. In the beginning, all labels that belong to the currently selected images are shown together with their ancestors and descendants in the hierarchy. 

It is possible to add new nodes to the hierarchy by selecting words from the recommendations and searching for a specific word in the dataset. The added nodes of words are colored black. Furthermore, one can draw a new edge from one node to another and remove it, by clicking on an edge; after each operation, the layout is recomputed. The addition of nodes and edges allows one to classify metadata into categories like themes or objects and also to connect metadata from different datasets. Edges that were drawn by other users are presented by a dashed line in the label hierarchy.

\textbf{Usage Scenario.}
We use the label hierarchy creation view to create child and parent relationships between related labels. Creating relationships to link metadata in both datasets works relatively well, especially for contemporary objects, such as flora, fauna, body parts, body postures, or furniture. More specific elements of medieval manuscripts such as animalistic, human, or hybrid decorative elements, or even narrative referents such as crucifixion, sleep, or glance of the eyes, are difficult, or sometimes impossible, to place. 

User 2, for example, started the visual analytics process not by labeling the images with missing labels but by working on the hierarchy. Starting with the hierarchy helped to understand the variety of labels from a holistic point of view and to understand how they relate to each other. Annotating the illuminations with a good idea of the existing vocabulary acquired by working on the hierarchy improved the quality of the labeling. 
Knowing the hierarchy system also helps to be more precise in the description and the labels used; for example, a lower label, more detailed, encompasses more information.

\begin{figure*}
 \centering
 \includegraphics[width=\linewidth]{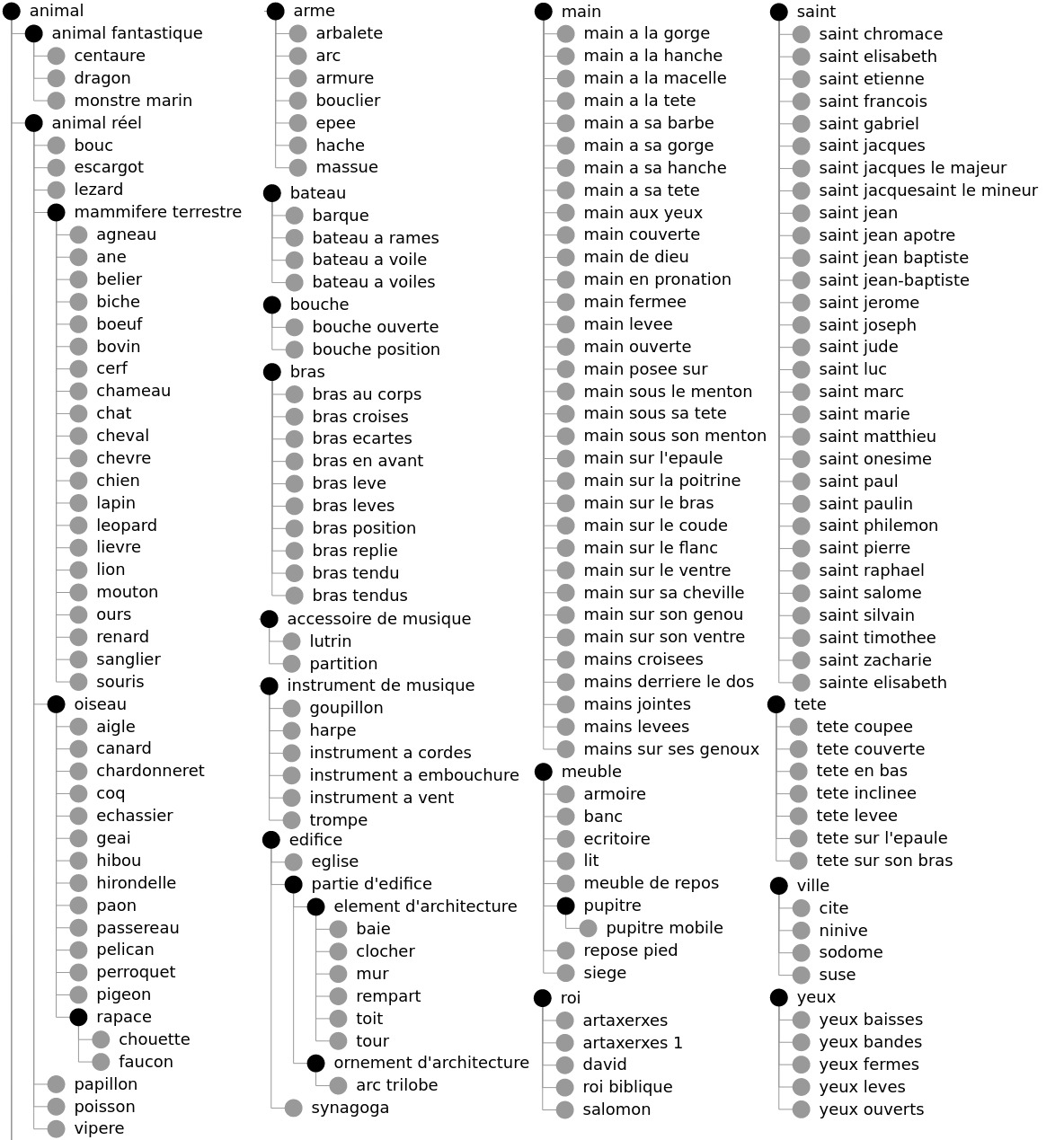}
 \caption{An excerpt of 201 nodes of the total 842 nodes that make up the current state of the label hierarchy of medieval Latin Bible illuminations. Leaves are colored grey, while inner nodes are black. The level in the tree is given by the indentation. The pseudo root of the tree is not displayed.}
 \label{fig:label}
\end{figure*}

\subsection{Feedback Computation}
New labels are added in real-time to the dataset, which also updates the label co-occurrences and can result in new recommendations in all views. In order to avoid unnecessary computations, the update of the graph similarities and the UMAP embeddings is done asynchronously in the background. It is also possible to see previous states of the graph and the embeddings using a slider.

The resulting label hierarchy (\autoref{fig:label}) is used to adjust word embeddings through retrofitting~\citep{faruqui:2015:NAACL}. This results in moving words closer together in the vector space that share an edge in the graph. For this, a second vector space is created and saved. This changes the nearest neighbor search of labels as now the union over the nearest neighbors of both vector spaces is presented in the word space.

\subsection{Unflattening Legacy Annotations}

\begin{figure*}[t]
 \centering
 \includegraphics[trim={0 1.5cm 0 1.5cm},clip,width=\linewidth]{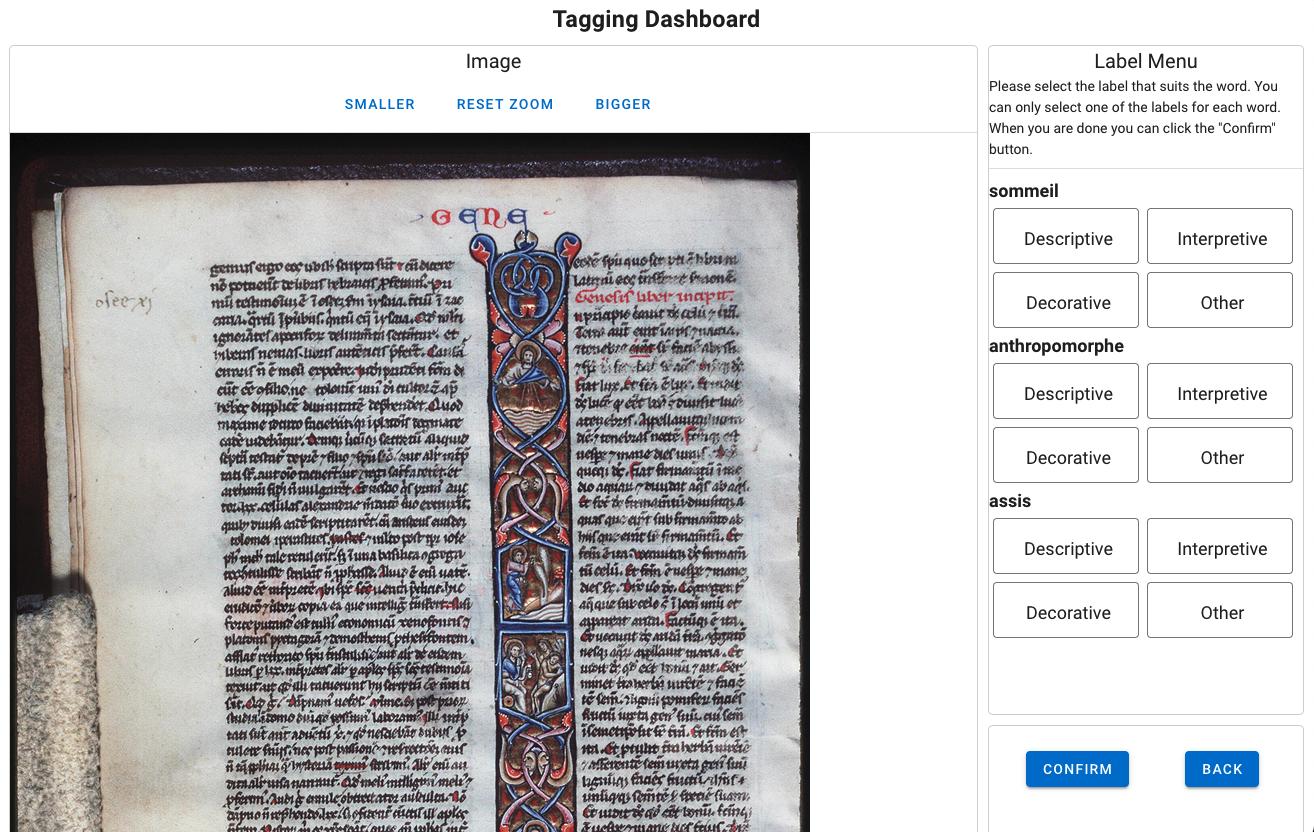}
 \caption{An example in our labeling dashboard of a manuscript illumination exhibiting all three of the types of labels: descriptive, interpretive and decorative. The manuscript depicted here is Bourges, Bibliothèque municipale 5, folio 3v.}
 \label{fig:tagging}
\end{figure*}

In our exploration of the legacy labels of the Mandragore and Initiale databases, we realized limitations for their use in working with computer vision for tasks such as object detection or image segmentation. The original labels had been designed for search and retrieval of images for forms of close viewing, rather than distant ones. Based on our thinking about hierarchies mentioned above in Section~\ref{s:5.4}, it occurred to us that we needed to subdivide the legacy annotations into categories of different orders of complexity that might be used for different downstream tasks. We unflatten the vocabularies by classifying them into three categories: "descriptive," "decorative" or "interpretive." For this, we designed a simple labeling dashboard that presents three of the possible labels for a given image from the dataset.

We qualify descriptive examples as those that anyone, without specialized knowledge, say a child, could learn to recognize. An illumination might contain "chauve" (bald), or an "épée" (sword) or a "barbe" (beard). In the case that the label is quite close to contemporary image hierarchies, it would be a somewhat straightforward task for computer vision. The case of \autoref{fig:tagging} "assis" (seated) is a body posture, a type of label that specialists in iconography take great interest in and that the database Initiale uses quite often~\citep{garnier1982langage}. Whereas perhaps not as straightforward as a beard or a sword, it is nonetheless easy to recognize without expert knowledge, and existing models perhaps provide a starting point~\citep{schlecht2011detecting}. The second category of labels is "decorative" and largely refers to a set of very detailed labels that only a manuscript specialist would be proficient in. These include motifs such as the "rinceau" (a wavy form depicting leafy stems), the "protomé" (a partial representation of a body found at the end of a decorative motif) or, as in \autoref{fig:tagging} "anthropomorphe" (a motif in the shape of a human figure). Theoretically, someone could learn to recognize these forms as well, although the vocabulary for them is highly specialized and their prevalence in the dataset can sometimes be uncommon. The third category of labels is "interpretive" by which we mean that some extra-visual context (usually textual context) is required to be able to label the item. Examples might include a proper name such as "Saint Jean" where no particular iconographic element points to the figure being John. Alternatively, a label of a book of the Bible may be given to a highly generic initial form, and only its placement in the codex provides the clue. Or, as in \autoref{fig:tagging}, "sommeil" (sleep) is present in Genesis 2:21 and may be indicated by the posture of a human figure, but can only be understood by association with the text in mind.

Categorizing these labels into different sorts is an extension of our visual thinking process discussed in Section~\ref{s:3}. Our assumption in doing so is that different sets of labels will facilitate different distant viewing goals. One example of this might be a distant thematics of manuscript illumination in which we look at the co-presence of specific objects. Another example using decorative elements might entail a kind of "stylometry of image" where we examine very specific motif patterns to identify particular illumination workshops. It is worth noting that the descriptive and decorative categories seem to be different orders of specificity, with the decorative being very specific to medieval art. That being said, the decorative could provide interesting pathways into cross-medium explorations in medieval art history. We are particularly interested in pursuing our visual thinking process with examples of the interpretive type, to understand not only how the legacy database labelers conceived of them, but also to explore how iconography communicates complex meanings. One initial hypothesis is that a combination of descriptive items found in an image can "arithmetically" add up to an interpretive judgment, but this instinctual reaction to our corpus needs to be verified empirically. The process of annotating illustrated to us that the thresholds between such categories of labels can be quite complex, making the choice of any label a complex one. In particular, when it comes to the hybridity of figures depicted in and the highly specific vocabulary used for medieval manuscripts, quite a lot of work needs to be done.

\section{Discussion}
Following a participatory design process~\citep{gappaper}, we extend Munzner's nested model~\citep{munzner2009nested} with frequent interdisciplinary exchange on a multitude of design aspects of our system. Thus, our system's visual interfaces were subjected to regular implicit evaluations by target users. This process revealed limitations, some of which were addressed during our iterative process, on the one hand, and future directions, on the other.

\subsection{Computational Limitations}
The current state of the label hierarchy includes 842 terms in the dataset and can be used for image classification tasks, for example, for a specific subtree of the hierarchy like "positions"~\citep{dh2023}. To further include object localization tasks, an additional labeling process would be needed to add bounding boxes, or pixel-wise masks. In addition, including other types of relationships outside of parent-child relationships, such as synonyms, could be of interest. Furthermore, including more means to show the data distribution could help in the process of creating the label hierarchy.

We did not extract the illuminations from the scanned manuscripts prior to processing because the state-of-the-art methods for doing so were not robust. The complexity of page structures, backgrounds, and preservation statuses led to insufficient and unusable results. Depending on the content of the raw image, this can lead to a smaller depiction of the illuminations and also in background noise through the text and the background. Another limitation lies in the cross-depiction problem~\citep{hall2015cross}, as the applied neural network to compute the image embeddings is pre-trained on natural images and not on medieval illuminations.

\subsection{Qualitative Evaluation}
The nature of the dataset makes a quantitative evaluation of the system at this stage of our research near impossible. However, we are able to offer a qualitative analysis of its implementation and a comparison of the observations of the two users across Section~\ref{s:5}. We have created a system that allows for labeling and hierarchy creation among the full dataset of over 13000 images and 2000 labels. In this first step of research with the combined datasets of Initiale and Mandragore, the two users working on the system were able to generate with relative ease an initial labeling hierarchy (\autoref{fig:label}). In most cases, parents were created imminently from existing labels and, in a minority of cases, they were created anew, based on abstraction. With more time, users are enthusiastic about progressing, as the process allows the inherent relationships of the labels to emerge in an organized manner. 

Furthermore, the two approaches they took to the materials in the system were neither contradictory nor inconsistent with each other's work. Instead, both users wanted the annotation space and the hierarchy builder to be linked so that their work was complementary. The success of the system in our eyes is its interconnected and non-contradictory qualities, which allow multiple users with different experiences of analytics systems, metadata, and humanities databases to work in a cohesive and favorable manner compatible with their experiences. It was commented that because the datasets already contain labels, the hierarchy will probably expand faster than the number of labels, which may actually be beneficial for passing to a new stage of the work including downstream tasks using it, such as automated object recognition.

\subsection{Future Works}
In addition to the limitations that point to refinement of the system, we see multiple pathways for future work. For both the hierarchy and labels, instead of the user cycling through the images manually to complete the tasks, the system can be combined with incremental machine learning methods that combine automatic recommendation with user interaction, helping to create robust and generalizable models. One possibility would be to include an active learning component to combine it with the current labeling process. Also, including methods to focus on the uncertainties of other metadata, like spatio-temporal annotation, could be a next step. Such a system moves from labeling and training to prediction, illustrating relationships that the improved metadata can suggest.

It would be interesting to extend the system in the form of an image labeling and exploration system that is conceived for several actors, takes input from them, and visualizes disagreement and debate amongst scholars about how we label and classify, but also attempts to provide multiple classifications. Given that our system is designed to create a label hierarchy as we annotate and there is potential disagreement between annotators, we will need to explore visual modes for representing (and resolving) ambiguity and disagreement. For labels, methods based on Fleiss’ Kappa~\citep{fleiss1971measuring} could be used to display inter-annotator agreement. Visualization should help to keep the human in the system, by showing the different paths users took and the different decisions they made, and so making the collective contribution of knowledge and labor visible.

Cultural collections of images are siloed, and metadata can be inconsistent or non-existent. The knowledge of the images found in this single genre (the Latin bible of the thirteenth and fourteenth centuries) should obviously be expanded to encompass many more periods and genres of Christian art. The larger picture of the project is to leverage what is known about the images in one place with some expert input to expand what is known in others. The hierarchy co-constructed in our visual analytics system would ideally be used to expand to other genres of manuscripts. In the long view, learning how to predict classification for the entire dataset or any other digitized collections of medieval art could be focused on repeated themes; style transfer to stained glass, or other cross-genre depictions found in other legacy art historical collections. Such a system would be of benefit to the larger community of medievalists and other scholars interested in images.

\section{Conclusion}
Carrying out a systematic relabeling of images from two legacy iconographic databases would be almost impossible by hand. Not only has it taken public institutions many decades to get to the point where the data are now, but the two sets are artificially siloed, as we have mentioned above. Both tasks--adjusting the labels based on recommendations and creating the hierarchy of labels--as organized in our visual analytics system provide a framework for understanding the logic of previous annotators, but also to rethink and expand the two datasets of common cultural artifacts into a more unified dataset. 

In this paper, we have presented a visual analytics process to create both annotations and label hierarchies based on a dataset of medieval illuminations. The system itself gives access to a large image dataset by providing different entry points to understand complex relations between manuscripts, images and the ways that these images have been studied by generations of art historical scholarship. We argue that such a system can be generalized to a number of different cultural heritage collections where metadata gaps prevent holistic discoverability. It supports the labeling process by combining machine learning methods with interactive visualizations. The resulting annotations and the label hierarchy are in a preliminary stage at present but can be iteratively refined and used for machine learning tasks where contemporary hierarchies are not appropriate. The results presented here are just a starting point to build bridges between the artificial siloes created in historical research and also to provide more complex, multi-faceted access to the illuminations in later work.

Furthermore, we presented cases of how subject specialists might would work with such a system, descriptions of what tasks were most appealing to them, a qualitative evaluation of the state of the research, and an assessment of the current limitations and potential for future work in this domain.


\bibliographystyle{agsm}

\bibliography{template}
\end{document}